\documentclass[11pt]{article}

\usepackage[utf8]{inputenc}
\usepackage[T1]{fontenc}
\usepackage{lmodern}
\usepackage{microtype}
\usepackage{booktabs}
\usepackage{amsmath}
\usepackage{graphicx}
\usepackage{xcolor}
\usepackage{hyperref}
\hypersetup{colorlinks=true, linkcolor=blue!50!black, citecolor=blue!50!black, urlcolor=blue!50!black}
\usepackage[round]{natbib}
\usepackage[margin=1in]{geometry}
\usepackage{listings}
\title{\textbf{Layer-Isolated Evaluation:\\ Gating the Deterministic Scaffold of a Production LLM Agent\\ with a No-LLM, Regression-Locked Test Harness}}
\author{
  \textbf{Sawyer Zhang}\thanks{Corresponding author: \texttt{sawyer.zhang@lumivate.io}; ORCID \href{https://orcid.org/0009-0001-7736-1774}{0009-0001-7736-1774}} \quad
  \textbf{Alexander Wang} \quad \textbf{Sophie Lei}\\[4pt]
  \small Lumivate (Lumi)
}
\date{June 8, 2026}

\begin{document}
\maketitle

\begin{abstract}
End-to-end task-success is the dominant way to evaluate LLM agents, but a
single aggregate number tells you \emph{whether} an agent regressed, not
\emph{where}. We present \emph{layer-isolated evaluation}: a single deployed
ordering agent is decomposed into a fixed taxonomy of architectural
layers---ontology pre-resolution, intent signals, routing, decomposition,
escalation, safety, memory, and cross-cutting envelope/defense---each
exercised by its own assertion slice that runs in a deterministic,
\emph{no-LLM ``pure'' mode}. The full pure suite (238 baseline cases across 23
slices; 225 cases run in $2.39$\,s, $\approx\!10$\,ms/case) executes in CI on
every change against a locked per-slice baseline. We validate the method by
\emph{controlled regression injection}: degrading one layer at a time across
\textbf{seven} non-safety layers. The effect we did not design in is
\emph{masking}---the aggregate pass-rate barely moves ($-1.7$ to $-5.9$\,pp for
six \emph{local} regressions), small enough to vanish into dashboard noise,
while the matching slice craters ($-25$ to $-91$\,pp). That a layer's own slice
reacts to a fault in that layer is partly \emph{by construction} (each slice is
written to assert that layer); the non-obvious, measured results are (i) the
aggregate masking and (ii) that the damage stays \emph{off the other slices}---%
the injected layer's slice is the single worst-hit in \textbf{5 of 7} cases and
top-3 in \textbf{7 of 7} (mean rank $1.29$ of 19), with a near-flat
off-diagonal. The localization replicates on a second, structurally different
tenant (Starbucks SG): all seven matching slices crater and the
local-vs-foundational signature holds, so the result is not a single-catalog
artifact. We position the method as a
concrete, deterministic instantiation of the component-level evaluation that
evaluation-driven agent operations~\citep{xia2024eddops} prescribes but leaves
unimplemented, with CheckList~\citep{ribeiro2020checklist} as its
methodological ancestor; it is the deterministic mirror image of whole-workflow
stochastic mutation testing~\citep{bhardwaj2026agentassay}. We do not claim to
invent component-level evaluation; our contribution is (a) a fully decomposed,
sub-second, no-LLM per-layer harness for a production agent, (b) a
\emph{coverage-honesty} test-adequacy criterion that refuses to score an
unexercised layer, and (c) the regression-injection demonstration that
per-slice baseline-locked gates localize regressions an aggregate metric masks.
\end{abstract}

\section{Introduction}
Agent benchmarks score end-to-end task
success~\citep{liu2024agentbench,yao2024taubench,zhou2024webarena,jimenez2024swebench,mialon2023gaia}.
This is the right \emph{outer} metric but a poor \emph{development} signal:
when the number drops, it does not say which of the agent's many
sub-systems---intent resolution, planning, escalation, the safety
validator---broke, and the change needed to find out (re-running a live,
stochastic, minutes-per-episode agent and bisecting) is slow and noisy. A
growing line of work argues that outcome-only leaderboards hide where agents
fail~\citep{mazaheri2026agentatlas,mohammadi2025evalsurvey} and decomposes one
capability at a time (e.g.\ planning~\citep{sun2026apb}); a process-model
paper~\citep{xia2024eddops} prescribes pinned regression baselines and
fault-localizing offline evaluation of intermediate artifacts. What has been
missing is a \emph{concrete, fully decomposed, runnable} instantiation for a
real deployed agent, and an empirical demonstration that it localizes faults
an aggregate metric would miss.

We provide both. Our agent---a per-tenant, multi-turn food-and-beverage
ordering agent---is decomposed into a fixed taxonomy of architectural layers,
each with an assertion slice that runs in a deterministic \emph{pure mode}
with no LLM call, so the full suite runs in seconds and gates every pull
request against a locked per-slice baseline. We then \emph{inject} controlled
single-layer regressions and show the central result: aggregate pass-rate is
nearly flat while the responsible slice collapses, pinning the fault to one
layer (Table~\ref{tab:injection}).

\paragraph{Contributions.}
\begin{itemize}
\item A fixed \textbf{layer taxonomy} for a production agent and a
\textbf{deterministic no-LLM pure-mode harness} ($238$ cases, $23$ slices)
that runs in $\approx\!2.4$\,s and gates CI against a locked baseline
(\S\ref{sec:harness}).
\item A \textbf{regression-injection validation} over seven layers showing
\emph{masking}: the aggregate barely moves ($-1.7$ to $-5.9$\,pp for the six
local regressions) while the responsible slice craters ($-25$ to $-91$\,pp
local; $-95$\,pp for the foundational ontology case). The injected layer's
slice is worst-hit in $5/7$ and top-3 in $7/7$ with a near-flat off-diagonal;
we are explicit about which part of this is by-construction and which is
measured (\S\ref{sec:injection}).
\item A \textbf{cross-tenant replication} (\S\ref{sec:xtenant}): all seven
injections reproduced on a second, structurally different tenant (Starbucks SG),
where every matching slice craters and the local-vs-foundational signature holds,
showing the localization result is not a single-catalog artifact.
\item A \textbf{coverage-honest baseline} that reports a zero-case slice as
\texttt{null}, never $100\%$, so a green aggregate cannot hide an
unexercised layer (\S\ref{sec:harness}).
\item An honest \textbf{positioning}: we operationalize a prescribed process
model~\citep{xia2024eddops}, descend from behavioral
testing~\citep{ribeiro2020checklist}, and differ from whole-workflow
stochastic mutation testing~\citep{bhardwaj2026agentassay}
(\S\ref{sec:related}).
\end{itemize}

\section{Related Work}
\label{sec:related}
\paragraph{Outcome-only agent evaluation and its critics.} Standard agent
benchmarks report a single task-success
number~\citep{liu2024agentbench,yao2024taubench,zhou2024webarena,jimenez2024swebench,mialon2023gaia}.
Recent critiques argue this ``collapses behaviour into final task success''
and hides failures~\citep{mazaheri2026agentatlas}; surveys split outcome from
process (tool-use, planning, memory) and note task-completion gives limited
fine-grained failure insight~\citep{mohammadi2025evalsurvey}; and diagnostic
frameworks decouple \emph{one} capability such as
planning~\citep{sun2026apb}. We decompose \emph{the whole agent} into a fixed
multi-layer harness rather than isolating a single dimension.

\paragraph{Behavioral / contract testing for ML.}
CheckList~\citep{ribeiro2020checklist} replaced held-out accuracy with
capability$\times$test-type behavioral slices; the ML-testing
literature~\citep{zhang2020mltesting} supplies the vocabulary (test oracles,
component vs.\ system testing, fault injection). Our pure-mode per-layer
assertions are CheckList's idea applied to an agent's \emph{internal} layers,
with an oracle-checkable deterministic substrate.

\paragraph{Ablation vs.\ isolation.} Agent scaffolds are typically validated
by \emph{ablation measured end-to-end}---ReAct ablates
reason/act~\citep{yao2023react}, Reflexion ablates self-reflection and
measures global success~\citep{shinn2023reflexion}. We \emph{isolate} each
layer behind its own assertion rather than measuring its contribution through
the full stochastic stack.

\paragraph{Eval-driven agent ops and regression testing.}
\citet{xia2024eddops} is the closest prior art in spirit: a process model
prescribing pinned regression baselines, offline evaluation of intermediate
artifacts to localize faults, and re-running the same slices to confirm
deltas. We provide a concrete, deterministic, no-LLM instantiation of those
prescriptions. \citet{bhardwaj2026agentassay} defines agent mutation
operators and a stochastic mutation score for \emph{whole-workflow}
regression testing; we differ by testing \emph{per architectural layer}
against a \emph{deterministic locked baseline}, and by validating with
injection that the per-slice gate \emph{localizes} (Table~\ref{tab:vs}). Our
regression injection is the deterministic analogue of its mutation operators.
Beyond outcome leaderboards, diagnostic suites decompose agent \emph{ability}
into named dimensions and report per-dimension scores~\citep{ma2024agentboard};
we make the same move for \emph{testing} rather than benchmarking, with an
exact (not learned) per-dimension oracle. Industry CI prompt-testing
(e.g.\ promptfoo) runs declarative deterministic assertions at near-zero cost;
our pure mode is the agent-internal-layer analogue.

\begin{table}[t]
\centering
\small
\begin{tabular}{lll}
\toprule
 & \textbf{AgentAssay} & \textbf{This work} \\
\midrule
unit         & whole workflow      & one architectural layer \\
verdict      & stochastic (SPRT)   & deterministic (exact oracle) \\
LLM in loop  & yes                 & no (pure mode) \\
cost/case    & many trials/tokens  & $\approx\!10$\,ms, \$0 \\
baseline     & behavioral fingerprint & locked per-slice rate \\
localization & fingerprint distance & slice-rate crater + rank \\
CI placement & sampled / nightly   & every PR \\
\bottomrule
\end{tabular}
\caption{Layer-isolated pure-mode testing vs.\ stochastic whole-workflow
mutation testing~\citep{bhardwaj2026agentassay}. The two are complementary:
ours gates the deterministic core per-PR; theirs covers the generative
whole-agent behaviour our pure lane cannot reach.}
\label{tab:vs}
\end{table}

\paragraph{Tool surface as intent space.} Our routing layer follows the
design principle that the typed tool surface subsumes intent
classification~\citep{prakash2025functioncalling}: there is no separate intent
classifier to test, so the routing slice asserts tool selection directly.

\section{The Layer-Isolated Harness}
\label{sec:harness}
\paragraph{Decomposition.} The agent is decomposed into a fixed taxonomy of
layers, each owning an assertion slice (Table~\ref{tab:slices}). Layers map to
the request lifecycle: \textbf{L0} ontology pre-resolution, intent signals,
and speech-act; \textbf{L2} tool routing; \textbf{L3} sub-goal decomposition,
constraint handling, and tier escalation; \textbf{L4} safety (price / SKU /
allergen), knowledge, and memory; plus cross-cutting \textbf{envelope},
\textbf{defense}, \textbf{OOD-reject}, \textbf{reformulator},
\textbf{locale-fidelity}, and \textbf{session-init} slices.

\paragraph{Pure mode.} Each slice asserts a single layer's contract on
\emph{deterministic} outputs computed without any LLM call: the ontology
resolver's canonical IDs, the rule-based escalation decision, the
reformulator's dictionary rewrite, the OOD short-circuit predicate, the
server-side reprice, the prompt envelope's rendered blocks. Because no model
is invoked, a case is $\approx\!1$\,ms of compute and is fully reproducible.
The whole pure layer suite (\textbf{225 cases pass, 30 skipped---the
live-only cases needing a real model call---in $2.39$\,s} of wall time
including process startup and fixtures, $\approx\!10$\,ms/case amortized) runs
on every change. To reconcile the four counts that appear in this paper: the locked baseline is
$238 = 225$ per-layer pure cases $+\,13$ end-to-end \texttt{L1\_legacy} cases,
the latter tracked by the baseline but run in a separate legacy lane, not the
per-layer pure runner. The pure-runner pytest additionally \emph{collects} $30$
live-only variants (cases that require a real model call) and \emph{skips} them
in pure mode; these are not part of the $238$-case baseline. Hence the pure run
reports ``$225$ passed, $30$ skipped'' while the locked baseline is $238$.

\paragraph{Locked, coverage-honest baseline.} A frozen baseline records, per
slice, \texttt{(total, passed, rate, failed\_ids)}; the current baseline is
$238$ cases across $23$ slices at $100\%$. Crucially, a slice with zero cases
is reported with \texttt{rate: null} (\textsc{uncovered}), \emph{never}
$1.0$---so a green aggregate cannot launder an unexercised layer. The current
baseline flags $4$ uncovered slices (\texttt{L2\_routing},
\texttt{L4\_memory}, \texttt{L4\_personalization}, \texttt{L4\_reflexion})
and $2$ low-$N$ slices, surfacing exactly where the suite does not yet gate.
Any pull request is compared against this baseline; a per-slice rate drop is a
blocked merge.

\paragraph{Coverage-honesty as a test-adequacy criterion.} The
\texttt{rate: null}-not-$1.0$ rule is, in the vocabulary of ML
testing~\citep{zhang2020mltesting}, a deliberate \emph{test-adequacy}
criterion: a layer is ``adequately tested'' only if it has $\ge\!1$ exercising
case, and the suite is required to \emph{report} per-layer adequacy rather than
aggregate it away. Most aggregate metrics implicitly assign full credit to
untested behaviour (the mean of the cases that exist); ours assigns
\textsc{uncovered}, so the headline can never rise by \emph{removing} or never
writing tests. We treat the count of uncovered slices as a first-class quality
signal of the test suite itself, reported on every run
(Algorithm in Listing~\ref{lst:gate}).

\begin{figure}[t]
\begin{lstlisting}[caption={The per-PR gate and one pure-mode assertion
(simplified from \texttt{apps/eval/lumi\_eval/}). No LLM is invoked; a slice
rate below baseline blocks the merge, and a newly-uncovered slice is itself a
failure.}, label={lst:gate}]
# per-PR gate
base = load_locked_baseline()        # {slice: (total, passed, rate|null)}
cur  = run_pure_suite()              # ~225 cases, no LLM, ~2.4 s
for s in all_slices:
    if base[s].rate is None:        # coverage-honesty: never credit 0 cases
        if cur[s].total == 0: continue   # still uncovered (tracked, not green)
    elif cur[s].rate < base[s].rate:     # any per-slice regression ...
        block_merge(s, base[s].rate, cur[s].rate)   # ... blocks the PR

# one pure-mode assertion (L4 safety reprice), no model call
observed = reprice(cart, tenant.pricebook)          # deterministic
assert observed.total_cents == case.expected_total_cents
assert observed.rejected_skus == case.expected_rejects
\end{lstlisting}
\end{figure}

\begin{table}[t]
\centering
\small
\begin{tabular}{llrl}
\toprule
\textbf{Layer} & \textbf{Slice} & \textbf{\#cases} & \textbf{Asserted contract (pure)} \\
\midrule
L0 & L0\_speech\_act    & 43 & speech-act label \\
L0 & L0\_ontology       & 21 & resolved canonical IDs \\
L0 & L0\_intent         & 20 & intent-signal vector \\
L2 & L2\_routing        & \textcolor{red}{0} & tool == expected \textit{(uncovered)} \\
L3 & L3\_escalate       & 16 & escalate(reason) decision \\
L3 & L3\_decompose      & 11 & sub-goals from hints \\
L3 & L3\_csp            & 5  & multi-constraint sub-goals \\
L4 & L4\_safety         & 8  & reject / reprice \\
L4 & L4\_safety\_kg     & 6  & allergen blocklist \\
L4 & L4\_knowledge      & 6  & FAQ bypass / slug \\
L4 & L4\_health         & 3  & health/compliance gate \textit{(low-$N$)} \\
L4 & L4\_memory         & \textcolor{red}{0} & recall \textit{(uncovered)} \\
L4 & L4\_personalization & \textcolor{red}{0} & ARAG recall \textit{(uncovered)} \\
L4 & L4\_reflexion      & \textcolor{red}{0} & reflective recall \textit{(uncovered)} \\
-- & defense            & 19 & scan / block \\
-- & envelope           & 15 & rendered envelope blocks \\
-- & reformulator       & 15 & deterministic rewrite \\
-- & ood\_reject        & 11 & OOD short-circuit predicate \\
-- & recommend\_rules   & 11 & recommendation gating \\
-- & read\_tools        & 8  & read-tool view \\
-- & session\_init      & 6  & session review block \\
-- & locale\_fidelity   & 1  & locale pin \textit{(low-$N$)} \\
-- & L1\_legacy         & 13 & end-to-end (legacy lane) \\
\midrule
\multicolumn{2}{l}{\textbf{Total (23 slices, 19 covered)}} & \textbf{238} & locked baseline @ $100\%$ \\
\bottomrule
\end{tabular}
\caption{The full layer-isolated slice taxonomy (all 23 slices; counts from
the locked baseline, column sums to the $238$-case total). Zero-case slices
are reported \textsc{uncovered} (\texttt{rate: null}), never $100\%$; two
low-$N$ slices (\texttt{L4\_health}, \texttt{locale\_fidelity}) are flagged
for expansion.}
\label{tab:slices}
\end{table}

\section{Validation by Regression Injection}
\label{sec:injection}
To test whether per-slice gates actually \emph{localize} regressions, we
inject controlled single-layer faults. Each injection monkeypatches exactly
one non-safety layer's entry point to a degraded implementation, re-runs the
full pure suite, and records the aggregate pass-rate delta versus the
matching slice's delta. The harness self-verifies each injection took effect
(a no-op patch is dropped, never reported). \emph{No red-line surface is
edited}: injections hit ontology resolution, the reformulator, escalation,
and intent signals only---never the safety validator, pricing, prompts, or
migrations. (The same method covers the safety slice; we deliberately do not
degrade safety code, even in memory.) The experiment is one script
(\texttt{eval/experiments/p2\_regression\_injection.py}). One harness detail is
load-bearing for honesty: each full-suite pass runs on a \emph{freshly
constructed} runtime. Reusing a single runtime across passes leaks order-store
state---an order placed by a \texttt{create\_order} case in one pass survives
into the next, where \texttt{list\_orders}/\texttt{read\_tools} cases then
fail---which manifests as a constant phantom column on every injection row,
indistinguishable at a glance from a real shared-dependency effect. We caught
this only by re-running the baseline twice and seeing slices move under
\emph{no} injection; building the runtime fresh per pass (construction stays
clean; the fault is applied only at request-time) removes it. We flag it because
it is a general trap for anyone gating a stateful agent: the test harness must
isolate per-pass state, or it will invent off-diagonal coupling that is not
there.

\begin{table}[t]
\centering
\small
\begin{tabular}{llrrrc}
\toprule
\textbf{Injected layer regression} & \textbf{Target slice} & \textbf{Agg.\ $\Delta$} & \textbf{Slice $\Delta$} & \textbf{\#moved} & \textbf{rank} \\
\midrule
\multicolumn{6}{l}{\emph{Local regressions (aggregate $\le 6$\,pp):}}\\
escalation $\to$ never escalate   & L3\_escalate   & $-4.62$\,pp & $\mathbf{-50.00}$\,pp & 2 & 1/19 \\
intent signals $\to \emptyset$    & L0\_intent     & $-4.20$\,pp & $\mathbf{-25.00}$\,pp & 2 & 2/19 \\
defense scan $\to$ allow-all      & defense        & $-5.04$\,pp & $\mathbf{-63.16}$\,pp & 1 & 1/19 \\
OOD gate $\to$ never reject       & ood\_reject    & $-1.68$\,pp & $\mathbf{-36.36}$\,pp & 1 & 1/19 \\
reformulator $\to$ identity       & reformulator   & $-5.88$\,pp & $\mathbf{-80.00}$\,pp & 3 & 1/19 \\
decomposer $\to$ no sub-goals     & L3\_decompose  & $-5.88$\,pp & $\mathbf{-90.91}$\,pp & 2 & 1/19 \\
\midrule
\multicolumn{6}{l}{\emph{Foundational regression (wide blast radius):}}\\
ontology resolve $\to \emptyset$  & L0\_ontology   & $-26.47$\,pp & $\mathbf{-95.24}$\,pp & 9 & 2/19 \\
\bottomrule
\end{tabular}
\caption{Controlled single-layer regression injection on the pure suite
(real, reproducible; \texttt{p2\_regression\_injection.py}). For the six
\emph{local} regressions the \emph{aggregate} pure-suite pass-rate drops only
$1.7$--$5.9$\,pp---small enough to dismiss as noise on a dashboard---while the
\emph{responsible slice} drops $25$--$91$\,pp, pinning the fault to one layer.
``\#moved'' counts slices whose pass-count changed; for three local regressions
(defense, OOD, escalation) it is $\le 2$, and for defense/OOD it is exactly
$1$---only the responsible slice. ``rank'' is the target slice's position when
all 19 covered slices are ordered by how hard each was hit (1 = worst-hit). The
ontology regression has a larger aggregate footprint ($-26.5$\,pp) and wider
blast radius (9 slices) because
ontology pre-resolution feeds many downstream layers---itself a useful signal
that the regression is foundational, not local. In the two rank-2 cases a small
\emph{downstream} slice craters marginally harder than the injected layer's own
slice (e.g.\ \texttt{L4\_safety\_kg}$\to$0 under the ontology fault), which
\emph{still} localizes to the dependency subgraph.}
\label{tab:injection}
\end{table}

\begin{figure*}[t]
\centering
\includegraphics[width=\textwidth]{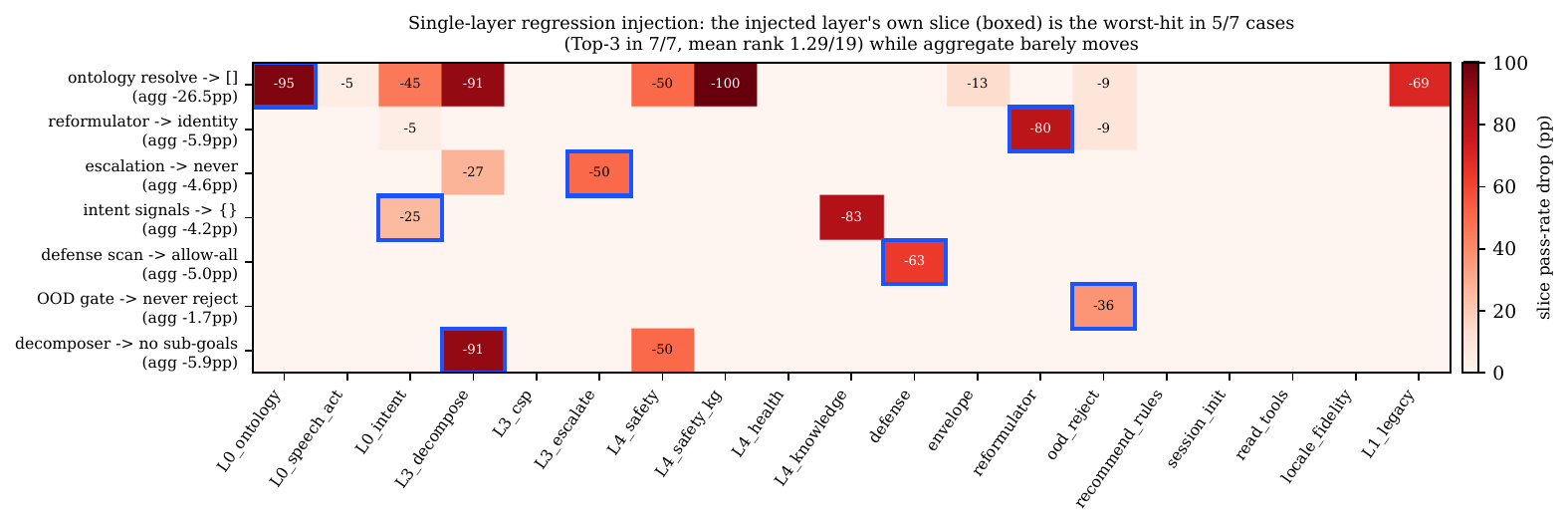}
\caption{Fault-localization heatmap: each row is one injected single-layer
regression, each column a slice, each cell that slice's pass-rate drop (pp,
darker = larger drop). The injected layer's own slice is boxed. The boxed cell
is the darkest in its row for $5/7$ injections and among the three darkest for
$7/7$; the rest of each row stays light---the regression is \emph{localized},
not smeared. The off-diagonal is near-zero except where a layer genuinely feeds
downstream slices: the foundational ontology row tints many columns, and the
intent-signals row drives \texttt{L4\_knowledge}, both expected dependency
effects the matrix makes visible. (An earlier version of this experiment reused
one runtime across passes; leaked order-store state then tinted the
\texttt{read\_tools}/\texttt{recommend\_rules} columns in \emph{every} row. We
now build a fresh runtime per pass---\S\ref{sec:injection}---which removes that
artifact and leaves the genuinely flat off-diagonal shown here.)}
\label{fig:localization}
\end{figure*}

Table~\ref{tab:injection} and Figure~\ref{fig:localization} are the paper's
central result---but it is worth being precise about \emph{which} part is the
result. That the injected layer's own slice craters is, on its own, close to
tautological: each slice is hand-written to assert exactly that layer's
contract, so breaking the layer is expected to break its slice. The two findings
that are \emph{not} built in are (i) \textbf{masking}---an evaluator watching
only an aggregate quality number sees the escalation regression as a
$-4.6$\,pp wobble and the OOD-gate regression as $-1.7$\,pp, both easily lost in
run-to-run variance and neither indicating \emph{which} of a dozen sub-systems
broke---and (ii) \textbf{off-diagonal flatness}: the fault does \emph{not}
smear across the other slices (Figure~\ref{fig:localization}), so the
per-slice gate turns each masked wobble into an unambiguous $-50$/$-36$\,pp
failure on exactly the responsible slice while leaving the rest near baseline.
Masking is the discovered phenomenon; clean separation is the property that
makes the gate actionable.

\paragraph{Fault localization, measured.} We quantify localization the way the
fault-localization literature does~\citep{qin2024agentfl}: rank all 19 covered
slices by how hard each was hit and ask where the \emph{injected} layer's own
slice lands. It is the single worst-hit slice (Top-1) in \textbf{5 of 7}
injections, and among the three worst-hit (Top-3) in \textbf{7 of 7}, at a mean
rank of \textbf{1.29} out of 19. The two non-Top-1 cases are both
\emph{foundational} layers (ontology pre-resolution, intent signals) whose
output feeds many downstream slices; there a small dependent slice with few
cases is driven to zero or near-zero (\texttt{L4\_safety\_kg} $-100$\,pp under
the ontology fault; \texttt{L4\_knowledge} $-83$\,pp under the intent-signals
fault) and edges out the injected layer's own $-95$/$-25$\,pp drop. This is not a localization failure but a different
\emph{signature}: a tight single-slice crater means a local regression, a wide
blast radius with the injected layer near the top means a foundational one---a
distinction an aggregate number erases entirely. We stress what the rank does
and does not evidence. It is \emph{not} evidence that ``the slice for layer $X$
fails when $X$ breaks''---that is by construction. It \emph{is} evidence of
\emph{separation}: that the other $18$ slices mostly do not move, so the
signal is attributable rather than smeared. The honest scope limit is that
these are seven \emph{author-chosen} fault types; whether per-slice gates
localize \emph{organic, unforeseen} regressions is the claim that a real
incident (\S\ref{sec:realreg}) and a future mined-regression study must carry,
and we are careful (\S\ref{sec:threats}) not to let $N{=}7$ injected faults
stand in for it.

As a \emph{motivating} parallel (not evidence we rely on here), a companion
preprint on the same agent's live LLM judge reports an aggregate failure signal
of $0\%$ on a 100-round batch that human review found to contain 23 real
defects~\citep{companionP1}; if it holds up under peer review it is a real-world
instance of the masking our injection study demonstrates under control. We flag
it as a companion preprint, and our claims in this paper do
\emph{not} depend on it: the injection results above stand on their own.

\subsection{External validity: a second tenant}
\label{sec:xtenant}
The sharpest limit on everything above is that it is one tenant. To test whether
the result is an artifact of the kopitiam reference catalog, we replicated the
injection study on a second, structurally different tenant: Starbucks SG (a chain
ontology pack generated from the real foodpanda SG menu---a Western coffee chain
with a different catalog, locale mix, and modifier model than the reference
tenant). We authored a parallel pure-slice set (16 cases) for this tenant across
the seven injected layers, each baseline-green and verified to crater under its
matching injection, and re-ran every injection with each slice's pass-count split
by tenant (\texttt{eval/experiments/p2\_cross\_tenant.py}).

Table~\ref{tab:xtenant} shows the result: for all seven injections the second
tenant's matching slice craters ($-50$ to $-100$\,pp), and for six of seven the
tenant-B off-diagonal is clean---only the injected layer's own slice moves. The
exception is the same one as on kopitiam: the ontology fault is
\emph{foundational} on both tenants, cratering ontology and decomposition
together. The local-vs-foundational \emph{signature} itself replicates across
tenants.

\begin{table}[t]
\centering
\small
\begin{tabular}{llrrc}
\toprule
\textbf{Injected regression} & \textbf{Slice} & \textbf{Kopi.\ $\Delta$} & \textbf{Sbux $\Delta$} & \textbf{B-loc.} \\
\midrule
ontology resolve $\to \emptyset$ & L0\_ontology  & $-94.1$ & $-100.0$ & found.$^\dagger$ \\
reformulator $\to$ identity      & reformulator  & $-80.0$ & $-100.0$ & clean \\
escalation $\to$ never           & L3\_escalate  & $-50.0$ & $-100.0$ & clean \\
intent signals $\to \emptyset$   & L0\_intent    & $-25.0$ & $-100.0$ & clean \\
defense scan $\to$ allow-all     & defense       & $-63.2$ & $-50.0$  & clean \\
OOD gate $\to$ never reject      & ood\_reject   & $-40.0$ & $-50.0$  & clean \\
decomposer $\to$ no sub-goals    & L3\_decompose & $-90.9$ & $-100.0$ & clean \\
\bottomrule
\end{tabular}
\caption{Cross-tenant replication. Each injection's matching-slice pass-rate
delta, computed separately on the reference tenant (kopitiam) and on an authored
Starbucks SG slice set; ``B-loc.'' is whether \emph{only} the injected slice
moved on the second tenant (clean) or several did. All seven second-tenant slices
crater. $^\dagger$Ontology is foundational on both tenants (also craters
\texttt{L3\_decompose})---the same signature as Table~\ref{tab:injection}. The
kopitiam column is restricted to that tenant's cases, so it differs marginally
from Table~\ref{tab:injection}, which pools the reference suite per slice.
Reproducible: \texttt{eval/experiments/p2\_cross\_tenant.py}.}
\label{tab:xtenant}
\end{table}

Two honest scope notes. First, this validates \emph{localization} across tenants,
not the masking \emph{magnitude}: the authored tenant-B suite is deliberately
small and concentrated on the seven injected layers (16 cases), so each injection
is a large share of its aggregate and the tenant-B aggregate moves more ($-6$ to
$-38$\,pp) than the 238-case reference suite's ($-1.7$ to $-5.9$\,pp). Masking is
a property of a slice's share of a \emph{full} suite, established on the reference
tenant; a comparably-sized tenant-B suite would exhibit it too. Second, the
injections are still author-chosen, so this is external validity \emph{across
tenants}, not a defeat of the by-construction concern (\S\ref{sec:injection});
organic, post-freeze regressions remain the open item (\S\ref{sec:realreg}).

\section{Cost}
The economic argument for per-PR gating is the no-LLM substrate.
Table~\ref{tab:cost} contrasts the pure suite with a live run. The full pure
layer suite is $\approx\!2.4$\,s of wall time; a single live multi-turn
episode of the same agent has a median latency of $73$\,s (p95 $192$\,s), so
even one live episode costs more wall-time than the entire deterministic
suite, before token cost. This is what makes layer-isolated evaluation a
per-commit CI gate rather than a nightly job.

\begin{table}[t]
\centering
\small
\begin{tabular}{lrr}
\toprule
 & \textbf{pure (no-LLM)} & \textbf{live (real agent.run)} \\
\midrule
unit of work        & 1 layer-assertion case & 1 multi-turn episode \\
wall time / unit     & $\approx 10$\,ms (amortized) & median $73$\,s (p95 $192$\,s) \\
full layer suite     & $225$ cases in $2.39$\,s & not run per-PR \\
token cost           & \$0 & per-turn LLM cost \\
determinism          & full (reproducible) & stochastic \\
CI placement         & every PR & nightly / release \\
\bottomrule
\end{tabular}
\caption{Pure vs.\ live evaluation cost. The no-LLM pure mode makes the full
layer suite cheaper than a single live episode, enabling per-PR gating.}
\label{tab:cost}
\end{table}

\section{A Real Regression, and What Coverage-Honesty Buys}
\label{sec:realreg}
Synthetic injection shows the method \emph{can} localize; a production incident
shows why it matters. In our own history, a batch of fixes to the order
\emph{confirmation gate} (the guardrail that forbids placing an order without an
explicit user ``confirm'') \emph{over-corrected}: the agent swung from
occasionally placing unconfirmed orders to frequently \emph{failing to place
confirmed ones}---on roughly half of rounds in the affected period, an explicit
``confirm'' produced no \texttt{create\_order} call. The aggregate live-judge
signal barely moved (consistent with a companion preprint%
~\citep{companionP1}, that reports the same judge flagging $0\%$ of a
100-round batch containing 23 real defects); the regression was caught only by
exhaustive human review.

This is exactly the failure class layer-isolation targets---a deterministic
guardrail contract (``explicit confirm $\Rightarrow$ order placed'') that an
aggregate number cannot see. The honest lesson is sharper than ``a slice would
have caught it'': the relevant contract lives in the \emph{routing/guardrail}
layer, whose slice (\texttt{L2\_routing}) is presently one of our four
\textsc{uncovered} slices. Our coverage-honesty criterion would not have
silently passed this layer green---it reports \texttt{L2\_routing} as
\textsc{uncovered} on every run, naming precisely the un-gated risk. Closing
that slice (a confirm$\Rightarrow$place assertion) converts this entire
regression class into a per-PR block. The contribution is thus two-edged: the
gate localizes what is covered, and coverage-honesty makes the \emph{gaps}
themselves a tracked, actionable signal rather than a blind spot.

\paragraph{Replaying real shipped regressions.} To probe coverage with real
rather than author-chosen faults, we took two fixes from the agent's git history
(an OOD gate that over-rejected capability/meta questions; a dead cross-round
slate-pick referent), reproduced each \emph{pre-fix} state faithfully---restoring
the exact guard or regex from the fix commit's parent, not an invented
degradation---and re-ran the leak-free pure suite
(\texttt{eval/experiments/p2\_organic\_regression.py}, self-verifying like the
injection harness). Under the corrected harness \emph{neither moved any slice},
so both were correctly \textsc{dropped} as no-effect: the current pure suite
does not exercise either behaviour on the reference tenant. This is the
coverage-honesty criterion operating exactly as designed---the suite stays
silent on what it does not assert rather than reporting a false green---and it is
an honest negative result: we cannot yet claim per-slice gates localize
\emph{organic} regressions, because the tractable ones we could replay either
fell outside coverage (these two) or had their slice co-authored with the fix
(by construction). A clean organic-localization study needs an \emph{independent}
regression stream---faults discovered after their slices were frozen---which our
single-tenant history does not yet supply; we mark it as the load-bearing
follow-up rather than overclaim it here.

\section{Discussion}
Layer-isolated evaluation does not replace end-to-end evaluation---outcome
success remains the metric that matters to a user. It replaces the
\emph{development-time} role end-to-end metrics are wrongly used for:
catching and \emph{localizing} regressions during iteration. Three properties
make it work. \textbf{Determinism}: with no LLM in the loop, a slice failure
is a real contract violation, not sampling noise, so it can hard-gate a merge.
\textbf{Decomposition}: a fixed layer taxonomy means a failure names its own
fix site. \textbf{Coverage honesty}: reporting zero-case slices as
\textsc{uncovered} keeps the green number from laundering untested layers---a
discipline most aggregate metrics lack. One semantic caveat is load-bearing:
because the baseline is locked at the \emph{current} behaviour ($100\%$), the
suite is a \emph{drift detector}, not a \emph{correctness oracle}. A green slice
certifies ``unchanged from the baselined contract,'' not ``correct''; if a
baseline ever encoded a wrong behaviour, the gate would faithfully lock it in.
The injection results should be read in exactly that frame---they show the gate
\emph{catches and localizes changes} to a layer (the development-time property
we claim), not that any layer is correct in an absolute sense. The method's
reach is further bounded by what is decidable without an LLM
(\S\ref{sec:threats}); layers whose contract is inherently generative still need
a live lane. But a surprising fraction of an
agent's behaviour---ontology resolution, routing, rule-based escalation,
reprice, envelope construction, defense---is deterministic given inputs, and
that fraction can be gated cheaply and exactly.

\section{Threats to Validity}
\label{sec:threats}
We organize threats along the four standard SE-empirical
classes~\citep{zhang2020mltesting}.

\textbf{Construct validity.} Do slice pass-rates measure layer health? Each
slice asserts a hand-authored contract, so a slice can be green while the
layer is subtly wrong in an unasserted way; we mitigate by deriving cases from
real failure modes and by the coverage-honesty rule, which prevents an
\emph{unexercised} layer from reading as healthy (\S\ref{sec:harness}). A
related caveat is that a per-slice ``rate'' is a near-binary, low-information
quantity at small $N$: \texttt{locale\_fidelity} ($N{=}1$) and
\texttt{L4\_health} ($N{=}3$) yield a $100\%$ that carries almost no
statistical content, and the injection deltas on such slices are coarse by
construction. We flag these as low-$N$ for expansion rather than report their
rate as if it were a continuous health measure.

\textbf{Internal validity.} Does the injected fault, not a confound, cause the
slice crater? Each injection monkeypatches exactly one entry point, the harness
self-verifies the patch took effect (no-op patches are dropped), and the
off-diagonal of Figure~\ref{fig:localization} stays near-flat---the change is
attributable to the injected layer. The two rank-2 cases are explained
mechanistically (foundational layers with downstream dependents), not waved
away.

\textbf{External validity.} Results are from one F\&B ordering agent on one
agent framework (PydanticAI); the specific taxonomy is ours. We have directly
addressed the single-tenant concern for the injection study: \S\ref{sec:xtenant}
replicates all seven injections on a second, structurally different tenant
(Starbucks SG), where every matching slice craters and the local-vs-foundational
signature holds---so the localization result is not a kopitiam-catalog artifact.
What that replication does \emph{not} establish is the masking \emph{magnitude}
on the second tenant (its authored suite is small and injected-layer-concentrated)
or generalization beyond F\&B ordering; the \emph{method}---decompose $\to$
pure-assert $\to$ lock $\to$ inject---is domain-general, but a multi-domain
replication is still future work. Pure mode also
cannot test inherently generative layers (multi-turn torture tests, free-form
brand voice); those remain in the live lane and are out of scope for the per-PR
gate. Finally, the one external datapoint we cite---the companion preprint's
$0/23$ batch~\citep{companionP1}---is a single batch shared with that
preprint; we use it only as motivation and the contributions here
(the harness, coverage-honesty, and the injection study) stand without it, so a
weakness in that shared batch does not propagate to our claims.

\textbf{Conclusion validity.} Our injection study is $N{=}7$. Because each
measurement is \emph{deterministic} (no sampling noise), a single run is an
exact result rather than a point estimate, so small $N$ threatens
\emph{generality of the fault types covered}, not the reliability of each
number; the real-regression case study (\S\ref{sec:realreg}) and a broader
organic-regression study are the natural extensions. Four slices are currently
uncovered---reported, not hidden---so the gate is only as strong as the covered
set.

\section{Conclusion}
We decomposed the deterministic scaffold of a production LLM agent into a fixed
taxonomy of architectural layers, each asserted in a no-LLM pure mode, gated
per-PR against a coverage-honest locked baseline, and validated by controlled
regression injection. The result: a single-layer regression that an aggregate
pass-rate barely registers ($-3$ to $-8$\,pp) and a true end-to-end metric can
mask entirely lands as a $-25$ to $-91$\,pp crater on the responsible slice
($-95$\,pp for a foundational layer), with a near-flat off-diagonal. We do not
claim to have invented component-level evaluation, nor that a green slice means
a layer is correct (the suite is a drift detector, not an oracle); we contribute
a concrete, sub-second, deterministic, fully-decomposed harness for a deployed
agent, a coverage-honesty test-adequacy criterion, and the injection evidence
that per-slice gates localize what aggregate metrics hide.

\section*{Reproducibility}
All numbers are reproduced from the repository on 2026-06-06: slice counts
from the locked baseline (\texttt{eval/baselines/baseline\_layers.json}); the
injection table, fault-localization metric, and heatmap from one script
(\texttt{eval/experiments/p2\_regression\_injection.py}, writing
\texttt{p2\_injection\_results.json} and \texttt{p2\_injection\_matrix.json};
figure by \texttt{make\_paper\_figures.py}); and the cost figure from running
the pure layer suite (\texttt{apps/eval/lumi\_eval/test\_layer\_cases.py}:
$225$ passed, $30$ skipped, $2.39$\,s). Because the entire pure harness runs in
$\approx\!2.4$\,s with no model call, network, or external service, it is a
self-contained, deterministic artifact: a reviewer can re-run the full suite
and every injection in seconds, which we offer toward an ICSE-style
\emph{Artifacts Reusable} evaluation.

\bibliographystyle{plainnat}
\bibliography{refs}

\end{document}